\pdfoutput=1
\def\year{2021}\relax
\documentclass[letterpaper]{article} 
\usepackage{aaai21}  
\usepackage[hyphens]{url}  
\usepackage{graphicx} 
\usepackage{natbib}  
\usepackage{caption} 
\usepackage{amsmath}

\usepackage[hidelinks]{hyperref}
\hypersetup{
	colorlinks=true,
	linkcolor=blue,
	filecolor=magenta,      
	urlcolor=cyan,
	bookmarks=true,
	citecolor=blue,
}
\usepackage{xspace}

\usepackage{cleveref}
\crefformat{footnote}{#2\footnotemark[#1]#3}

\usepackage{mathrsfs}
\usepackage[cal=boondoxo,scr=rsfso]{mathalfa}

\usepackage{nicefrac}

\newcommand{\com}[1]{}

\newcommand{\eq}[1]{(\ref{#1})}

\newcommand{\python}{\textsf{Python}\xspace}
\newcommand{\sympy}{\textsf{SymPy}\xspace}
\newcommand{\pytorch}{\textsf{PyTorch}\xspace}

\newcommand{\modelica}{\textsf{Modelica}\xspace}
\newcommand{\spice}{\textsf{Spice}\xspace}
\newcommand{\pdefind}{\textsf{PDE-FIND}\xspace}

\newcommand{\cyphy}{\textsf{\textbf{CyPhy}}\xspace}
\newcommand{\inet}{\textsf{\textbf{I-net}}\xspace}

\newcommand{\dimm}{\mathrm{d}}
\newcommand{\Dimm}{\mathscr{D}}
\newcommand{\ff}{\mathcal{f}}
\newcommand{\cc}{\mathcal{c}}
\newcommand{\dual}{\ast}
\newcommand{\loss}{\textsf{Loss}}

\title{AI Research Associate for Early-Stage Scientific Discovery}
\author{
	\rm
	Morad Behandish\textsuperscript{\rm 1}\thanks{Corresponding Author, e-mail: \href{mailto:moradbeh@parc.com}{moradbeh@parc.com}.},
	John T. Maxwell III\textsuperscript{\rm 1},
	Johan de Kleer\textsuperscript{\rm 1}
	\\
}
\affiliations{
	\textsuperscript{\rm 1}Palo Alto Research Center (PARC)\\
	
	3333 Coyote Hill Road, 
	Palo Alto, California 94304 (\href{http://www.parc.com/}{www.parc.com}) \\
	
}

\begin{document}

\maketitle

\begin{abstract}

Artificial intelligence (AI) has been increasingly applied in scientific activities for decades; however, it is still far from an insightful and trustworthy collaborator in the scientific process. Most existing AI methods are either too simplistic to be useful in real problems faced by scientists or too domain-specialized (even dogmatized), stifling transformative discoveries or paradigm shifts. We present an AI research associate for early-stage scientific discovery based on (a) a novel minimally-biased ontology for physics-based modeling that is context-aware, interpretable, and generalizable across classical and relativistic physics; (b) automatic search for viable and parsimonious hypotheses, represented at a high-level (via domain-agnostic constructs) with built-in invariants, e.g., postulated forms of conservation principles implied by a presupposed spacetime topology; and (c) automatic compilation of the enumerated hypotheses to domain-specific, interpretable, and trainable/testable tensor-based computation graphs to learn phenomenological relations, e.g., constitutive or material laws, from sparse (and possibly noisy) data sets.
%


\end{abstract}

\section{Introduction}

\noindent
Data-driven AI methods have been applied extensively in the past few decades to distill nontrivial physics-based insights (scientific discovery) and to predict complex dynamical behavior (scientific simulation) \cite{Stevens2020AI}.
Notwithstanding their effectiveness and efficiency in classification, regression, and forecasting tasks, statistical learning methods can hardly ever evaluate the soundness of a function fit, explain the reasons behind observed correlations, or provide sufficiently strong guarantees to replace parsimonious and explainable scientific expressions such as differential equations (DE). Hybrid methods such as constructing ``physics-informed/inspired/guided'' architectures for neural nets and loss functions that penalize both predication and DE residual errors \cite{Raissi2019physics,Wei2019physics,Daw2020physics} and graph-nets based on control theory and combinatorial structures \cite{Cranmer2019learning,Seo2019differentiable,Sanchez2020learning} are all important steps towards explainability; however, the built-in ontological biases in most machine learning (ML) frameworks prevent them from {\it thinking outside the box} to discover not only the known-unknowns, but also unknown-unknowns, during early stages of the scientific process. 

\com{
\begin{figure} 
	\centering
	\includegraphics[width=0.48\textwidth]{fig/fig_concept}
	\caption{The \cyphy framework}
	\label{fig2} \label{fig_concept}
\end{figure}
}

\subsection{Contributions} \label{sec_cont}

We present `cyber-physicist' (\cyphy), our novel AI research associate for early-stage scientific process of hypothesis generation and initial (in)validation, grounded in the most invariable mathematical foundations of classical and relativistic physics.
Our framework distinguishes itself from existing rule-based reasoning, statistical learning, and hybrid AI methods by:
\begin{enumerate}
	\item [(1)] an ability to rapidly enumerate and test a diverse set of mathematically sound and parsimonious physical hypotheses, starting from a few basic assumptions on the embedding spacetime topology; 
	\item [(2)]	a distinction between non-negotiable mathematical truism (e.g., conservation laws or symmetries), that are directly implied by properties of spacetime, and phenomenological relations (e.g., constitutive laws), whose characterization relies indisputably on empirical observation, justifying targeted use of data-driven methods (e.g., ML or polynomial regression); and
	\item [(3)] a ``simple-first'' strategy (following Occam's razor) to search for new hypotheses  by incrementally introducing latent variables that are expected to exist based on topological foundations of physics.%
	%
\end{enumerate}

\subsection{Background} \label{sec_lit}

AI-assisted discovery of scientific knowledge has been an active area of research \cite{Langley1998computer} long before the rise of GPU-accelerated deep learning (DL). 
As computational power and data sources are becoming more ubiquitous, model-based, data-driven, and hybrid AI methods are playing an increasingly more important role in various scientific activities \cite{Kitano2016artificial,Raghu2020survey}.

Related efforts to our approach to scientific hypothesis generation and evaluation are mostly engineered after how humans approach scientific discovery, including sequential rule-based symbolic regression \cite{Schmidt2009distilling,Udrescu2020AI}, latent space representation learning via deep neural net auto-encoders \cite{Iten2020discovering,Nautrup2020operationally} and strategic combinations of divide-and-conquer, unsupervised learning, simplification by penalizing description lengths in the loss function, and a posteriori unification by clustering \cite{Wu2019toward}. While these and other efforts have shown great promise for elevating AI to the role of an autonomous, creative, and insightful collaborator that can offer human scientists a set of viable options to consider, their applications have remained limited to rather basic examples. 

On the more domain-specialized end, DL has been widely successful in classification, regression, and forecasting tasks in scientific areas as diverse as turbulence \cite{Miyanawala2017efficient,Wang2020towards}, chaotic particle dynamics \cite{Breen2020newton}, molecular chemistry and materials science \cite{Butler2018machine}, and protein engineering \cite{Yang2019machine}, among others. Most specialized DL architectures are ad hoc, designed (by humans) using narrow, domain-specific, and (by construction) biased knowledge and expertise, stifling innovation and surprise.
Moreover, DL models that successfully capture nontrivial patterns in data are often difficult to explain, lack guarantees even within their training space, and poorly extrapolate to out-of-training scenarios \cite{Mehta2019high}. Training such models for high-dimensional physics problems requires enormous data, which is either unavailable or too costly to obtain in many experimental sciences. 

\section{The Cyber-Physicist}

We introduce an AI tool that can bridge multiple levels of abstraction, using a {\it domain-agnostic} representation scheme to express a wide range of mathematically viable physical hypotheses by exploiting common structural {\it invariants} across physics.
Our approach entails: 
\begin{enumerate}
	\item [(a)]	defining a relatively unbiased ontology rooted in fundamental abstractions that are common to all known theories of classical and relativistic physics;
	\item [(b)] constructing a constrained search space to enumerate viable hypotheses with postulated invariants, e.g., built-in conservation laws that are consistent with the presupposed spacetime topology; and
	\item [(c)] automatically assembling interpretable ML architectures for each hypothesis, to estimate parameters for phenomenological relations from empirical data.
\end{enumerate}
At the core of (a) is a powerful mathematical abstraction of physical governing equations rooted in algebraic topology and differential geometry \cite{Frankel2011geometry}. This abstraction leads to an ontological commitment to the relationship between physical measurement and basic properties of the embedding spacetime---but nothing more, to leave room for innovation and surprise. This relationship has been shown  to be responsible for the {\it analogies} and {\it common structure} across physics \cite{Tonti2013mathematical}, exploited in (b), along with search heuristics based on analogical reasoning. Each viable hypothesis is {\it automatically} compiled to an interpretable ``computation graph''---tensor-based architecture, akin to a neural net with convolution layers to compute differentiation/integration and (non)linear local operators for constitutive equations---for a given cellular decomposition of embedding spacetime using well-established concepts from cellular homology \cite{Hatcher2001algebraic} and exterior calculus of differential and discrete forms \cite{Bott1982differential,Hirani2003discrete} that are under-utilized in AI. 

\subsection{Topological Foundations of Physics}

The key enabler of our AI framework is a simple type system for (a) physical {\it variables}, based on how they are measured in spacetime; and (b) physical {\it relations}, based on their (topological vs. metric) nature, and the variables they connect. Following the ground-breaking discoveries by a number of mathematicians, physicists, and electrical engineers  \cite{Kron1963diakoptics,Roth1955application,Branin1966algebraic} towards a general network theory, Tonti explained the fascinating analogies across classical and relativistic physics in his pioneering life-long work \cite{Tonti2013mathematical} by reframing them in the language of cellular homology, leading to informal classification diagrams. %
%
%
Tonti diagrams can be formalized as directed graphs with strongly typed nodes for variables and edges for relations. The variable are typed as $(\dimm_1, \dimm_2)-$forms based on their measurement on $\dimm_1-$ and $\dimm_2-$dimensional submanifolds ($\dimm_1-$ and $\dimm_2-$cells) of space and time, respectively. For instance, to model heat transfer in (3+1)D spacetime, temperature is typed as a $(0, 1)-$form because it is measured at spatial points ($0-$cells) and during temporal intervals ($1-$cells), whereas heat flux is typed as a $(2, 1)-$form because it is measured over spatial surfaces ($2-$cells) and during temporal intervals ($1-$cells). In classical calculus, both of these variables reduce to scalar and vector fields, probed at spatial points and at temporal instants, to write down compact pointwise DEs; however, keeping track of the {\it topological and geometric character} of DEs is key to a deeper understanding of how known physical theories work, and building on top of it for AI-assisted discovery of new physics grounded in mathematical foundations.

\begin{figure*} 
	\centering
	\includegraphics[width=0.96\textwidth]{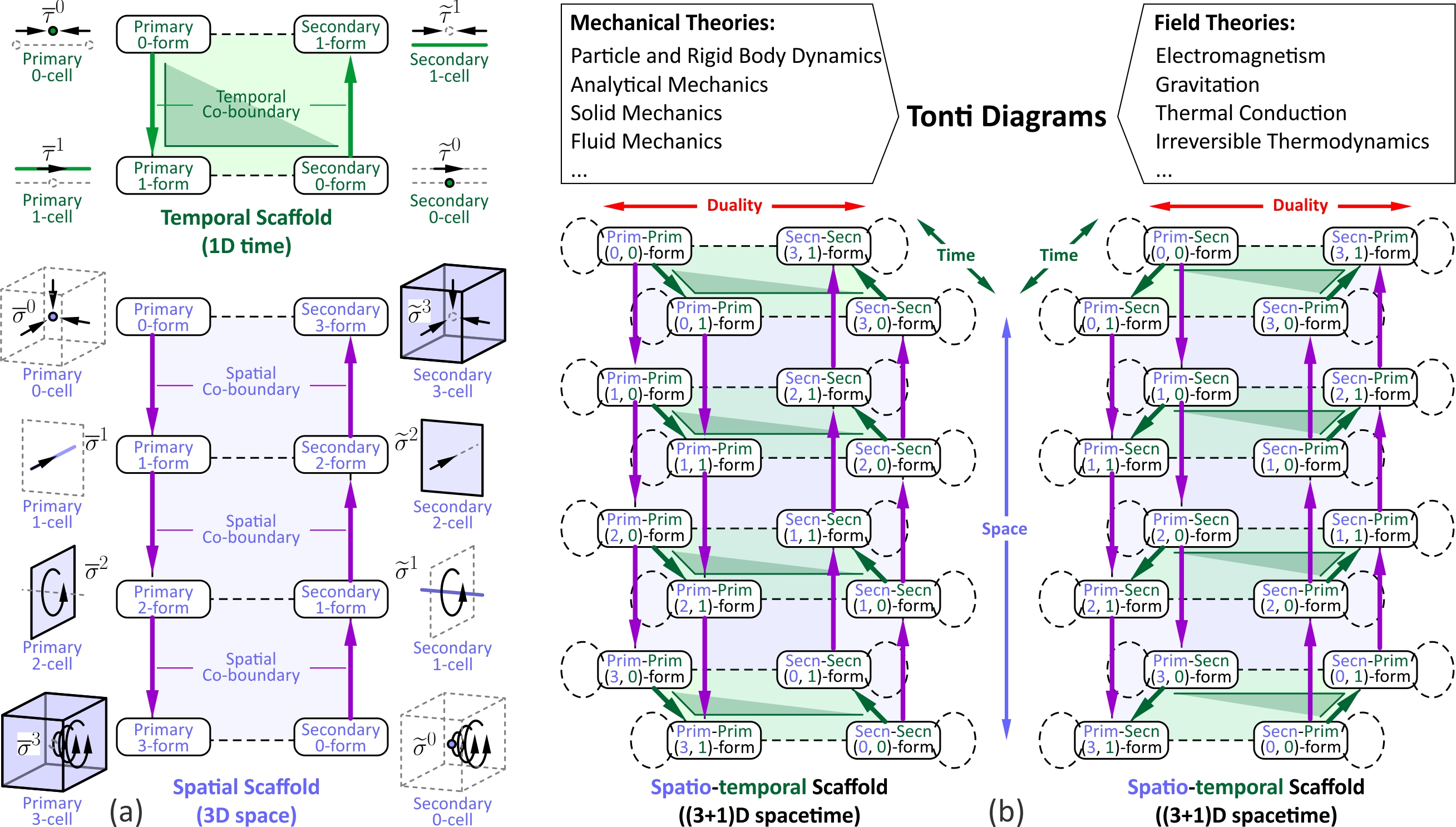}
	\caption{A topology-aware representation for physics \cite{Tonti2013mathematical}: (a) variables associated with spatial and temporal cells of various dimensions give rise to primary forms and secondary forms (also called pseudo-forms); (b) resulting in 32 possible types for spatio-temporal forms, and an underlying structure for fundamental theories of physics.} \label{fig_types}
\end{figure*}

The spatiotemporal cells (or embedding manifolds) are further classified as primary or secondary, endowed with inner or outer orientations, respectively, depending on how the variables change sign in a hypothetical reversal of spacetime orientation \cite{Mattiussi2000finite}. The cells are related by topological {\it duality} (Fig. \ref{fig_types} (a)). For example, an inner-oriented curve ($1-$cell, $\overline{\sigma}^1$) sitting in primary space, along which temperature variations are measured, is dual to an outer-oriented surface ($2-$cell, $\widetilde{\sigma}^2$) sitting in secondary space, over which heat flux is measured, and the two cells are spatially registered and consistently oriented, if we embed them in two co-located ``copies'' of 3D space.
%

\com{
\begin{figure} 
	\centering
	\includegraphics[width=0.48\textwidth]{fig/fig_heat}
	\caption{The heat equation in (2+1)D spacetime. In the infinitesimal length/time scale limit, the abstract equation reduces to the familiar PDE, but it could be interpreted differently in a discrete setting (Fig. \ref{fig_context}).} \label{fig_heat}
\end{figure}
}

\begin{figure*} 
	\centering
	\includegraphics[width=0.96\textwidth]{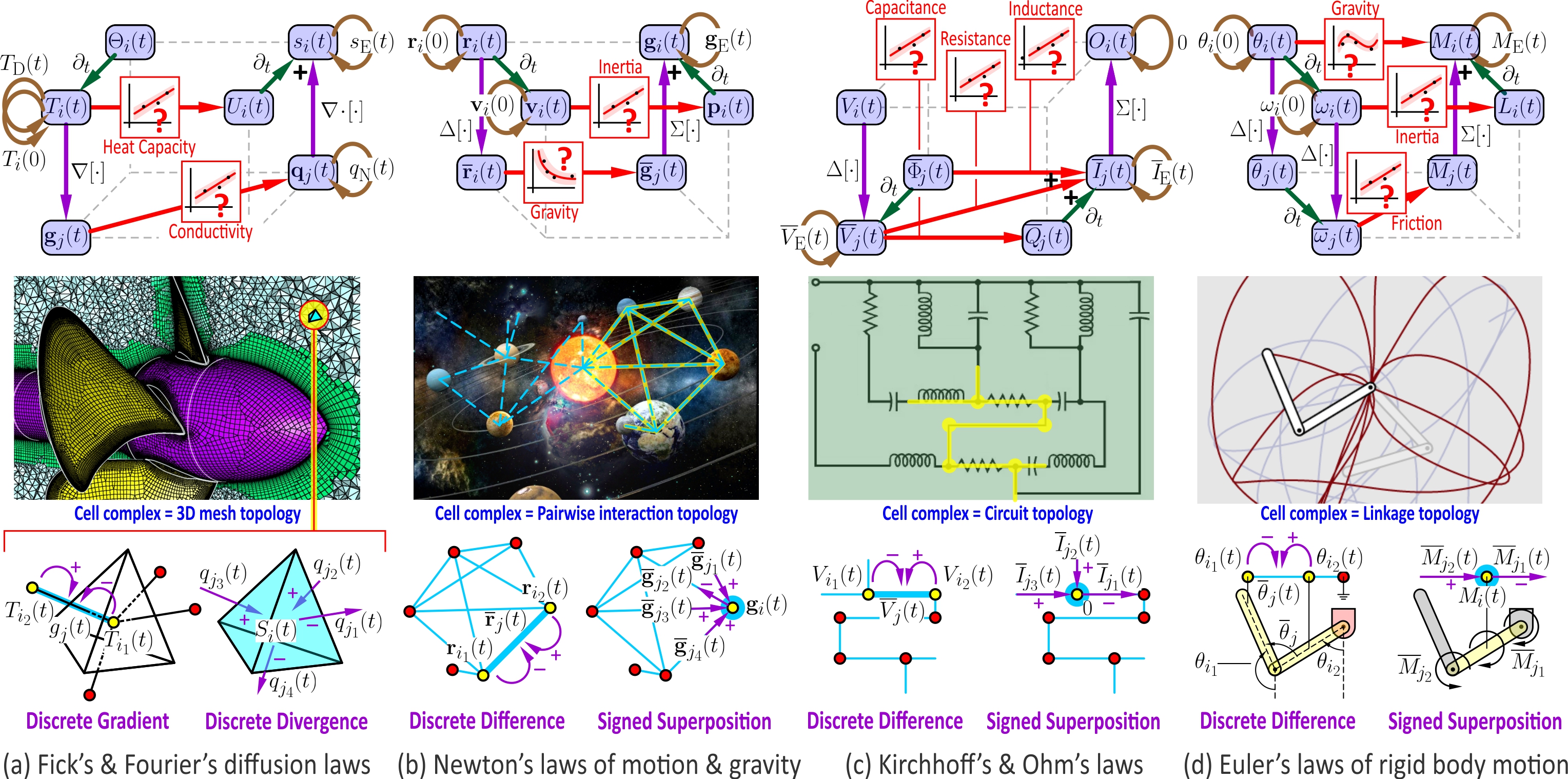}
	\caption{Tonti diagrams are recipes to generate governing equations in different contexts, defined by a continuum, discrete, or semi-discrete setting and a topological embedding of the variables based on how they are measured. The conservation laws in terms of co-boundary operators result directly from assumed properties of space (or spacetime), while constitutive relations must be learned from data (e.g., via regression/ML).} \label{fig_context}
\end{figure*}

The relations among variables on the Tonti diagrams are typed based on the pairs of variables they relate, as well as the nature of the relation itself:
\begin{itemize}
	\item {\it Topological} relations map spatiotemporal forms to forms of one higher dimension in space or time via incidence relations, and are responsible for propagation of information in spacetime through incident cells.
	\item {\it Metric} relations locally map forms defined over dual cells to one another based on phenomenological properties, spatial lengths, and temporal durations, and are responsible for local distortion of information.
	\item {\it Algebraic} relations are in-place, i.e., map a given form to another from of the same type, and can be used to capture initial/boundary conditions, external source/sink terms, or cross-physics couplings between variables of the same type on different diagrams.
\end{itemize}
The relations are drawn in Fig. \ref{fig_types} as vertical arrows, horizontal (or horizontal-diagonal) arrows, and loops ($1-$cycles), respectively. 
The {\it interpretation} of these relations to symbolic or numerical operations depends on the choice of a cellular decomposition of spacetime on which they operate. For example, using a continuum spacetime with infinitesimal cells, the variables are viewed as {\it differential forms} and the topological operators on them are interpreted as {\it exterior derivatives} \cite{Bott1982differential}. In elementary calculus, these operators give rise to gradient, curl, and divergence in space and partial derivative in time in terms of {\it scalar and vector fields} that are proxy to these forms, leading to partial DEs (PDEs). 
In a discrete (or semi-discrete) setting, on the other hand, the same diagram can be used to produce integral (or integro-differential) equations that capture the same fundamental conservation and constitutive realities, where the variables are viewed as {\it co-chains}, also called {\it discrete forms} (or mixed forms, e.g., discrete in space, differential in time, or vice versa) and the topological operators become {\it co-boundary operators} that are fundamental in cellular homology \cite{Hatcher2001algebraic}. For example, using a semi-discretization in space with integral quantities associated with $0-$, $1-$, $2-$ and $3-$cells on a pair of staggered unstructured meshes in 3D, while keeping time as a continuum, the semi-discrete form of the heat equation as a system of ordinary DEs (ODEs) (Fig. \ref{fig_context} (a)). Upon discretization of time, one obtains algebraic equations that can be solved or parameter estimated via tensor-based ML. 

\begin{figure*} [t]
	\centering
	\includegraphics[width=0.96\textwidth]{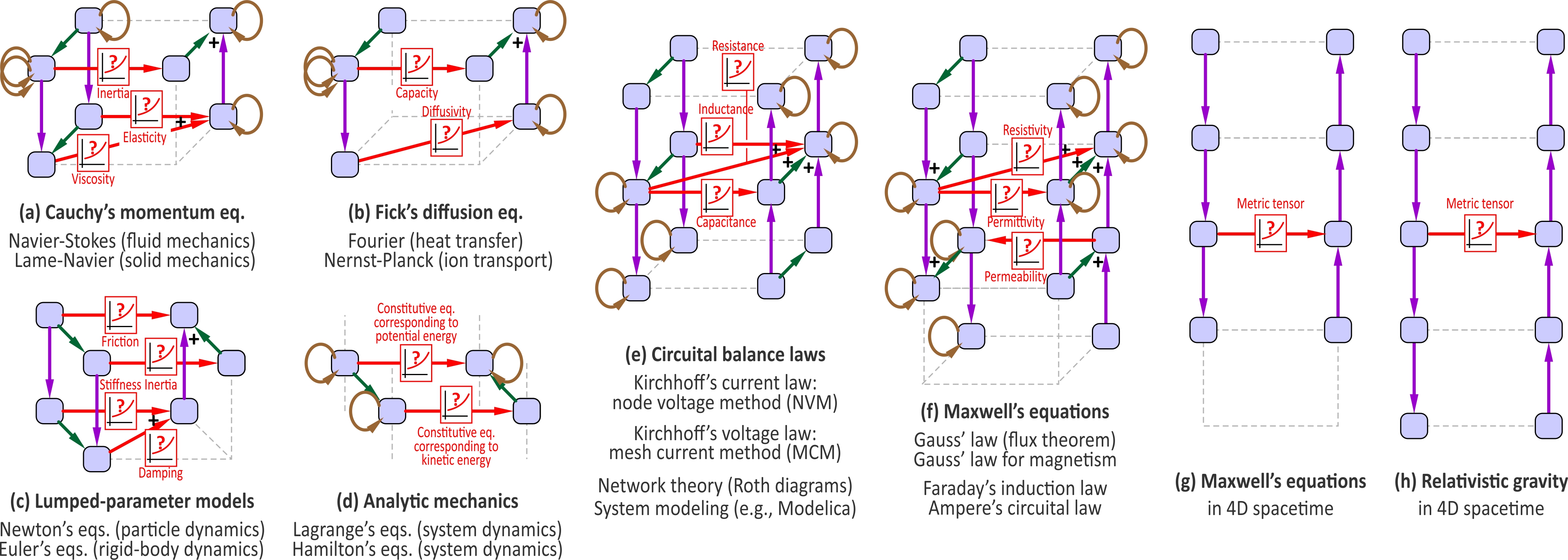}
	\caption{Tonti diagrams capture the common structure responsible for analogies across classical and relativistic physics with a clear distinction between topological and phenomenological relations that follow certain rules.} \label{fig_analogy}
\end{figure*}

It is important to note that 3D meshes in space and 1D time-stepping are not the only ways to provide a combinatorial topology to interpret Tonti diagrams in a discrete setting. Another example is a directed graph representation of lumped-parameter networks such as system models in \modelica or electrical circuits in \spice. The variables in this case are associated with nodes, edges, and meshes (i.e., primitive cycles) and incidence relations are obtained from graph connectivity and edge directions. The same topological operator that leads to a spatial divergence, discretized by a sum of fluxes on the incident faces of a volume in a 3D mesh, also leads to a superposition of forces on interacting planets, sum of currents in/out of junctions in electrical circuits, and superposition of torques on kinematic chains (Fig. \ref{fig_context} (b, c, d)). Both ODEs and PDEs and their integral or integro-differential forms upon full or semi-discretization can be captured with the same (abstract) operators, and Tonti diagrams serve as {\it recipes} to compose them to generate governing equations.

Figure \ref{fig_analogy} shows a few other examples of Tonti diagrams for fundamental theories in classical and relativistic physics. The differences amount to (a) topological and metric context; (b) relevant variables and their dimensions/units; and (c) phenomenological relations.

\begin{figure*} [t]
	\centering
	\includegraphics[width=0.96\textwidth]{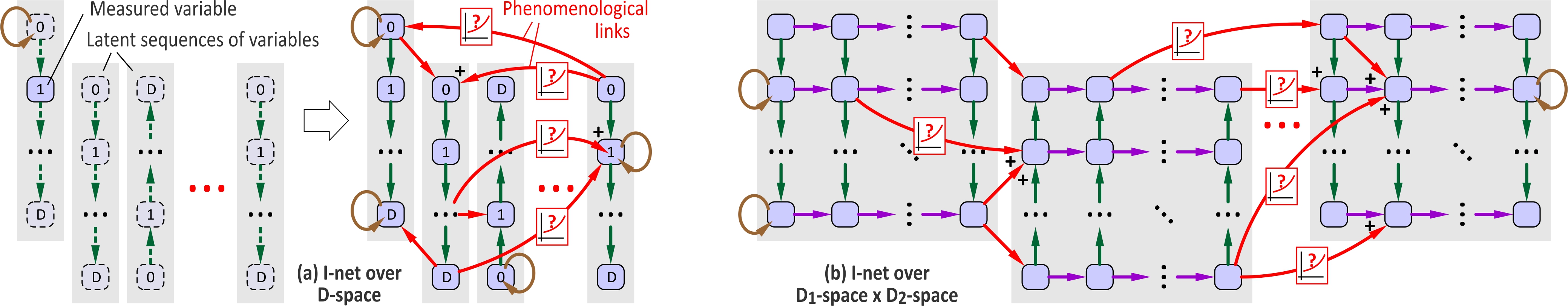}
	\caption{\inet{s} are generalizations of Tonti diagrams for finite topological products of finite-dimensional spaces with relaxed rules for feasible phenomenological links to accomodate middle-ground theories.} \label{fig_inets}
\end{figure*}

\subsection{An Ontology for Scientific Process}

We present a novel representation, called `interaction networks' (\inet{s}), based on a generalization of Tonti diagrams that is expressive and versatile enough to accommodate novel scientific hypotheses, while retaining a basic commitment to philosophical principles such as parsimony (Occam's razor), measurement-driven classification of variables, and separation of non-negotiable mathematical properties of spacetime (homology) from domain-specific empirical knowledge (phenomenology). Data science is employed to help only with the latter.
\begin{itemize}
	\item We conceptualize three levels of abstraction related by inheritance: abstract (symbolic) \inet{s} $\to$ discrete (cellular) \inet{s} $\to$ numerical (tensor-based) \inet{s}.
	\item At each level, an \inet instance is contextualized by user-defined assumptions on spacetime topology, semantics of physical quantities, and structural restrictions on allowable diagrams based on analogical reasoning and domain-specific insight (if available).
	\item Every \inet instance distinguishes between topological and metric operators; however, it has additional degrees of freedom (beyond Tonti diagrams) for the latter to allow for phenomenological relations among variables that may not be dual to each other.
\end{itemize}
%

The latter is motivated by the observation that some existing middle-ground theories use phenomenological relations to capture a combination of topological and metric aspects.
%

\com{
For example, Kepler's laws of planetary motion can be expressed as empirical constraints among relative positions and velocities of moving planets, which, despite being unfit for standard Tonti diagrams, can be expressed by simple \inet{} instances. 
Further dissection of these ``non-standard'' phenomenological relations can lead to Newton's laws of motion and gravity, offering a cleaner separation of topological and metric components.
The inverse-square law of gravity itself can be further explained by a combination of topological and metric parts using Poisson's formulation in terms of a potential field. 
Although our current implementation of AI search cannot navigate these transitions, the underlying representation scheme supports adding these capabilities in future.
}

We define an abstract (symbolic) \inet on a single $\Dimm-$space as a finite collection of primary and/or secondary co-chain complexes that are inter-connected by phenomenological links, as shown in Fig. \ref{fig_inets} (a). Each co-chain complex is a sequence of (symbolic) $\dimm-$forms related by (symbolic) co-boundary operators from $\dimm-$forms to $(\dimm+1)-$forms ($0 \leq \dimm \leq \Dimm$). The interpretation of $\dimm \to (\dimm+1)$ maps depends on the embedding dimension $\Dimm$; for instance, if $\Dimm = 1$ the only option for the input is $\dimm = 0$ leading to a simple partial derivative ($0 \to 1$), whereas for $\Dimm = 3$, we can have $\dimm = 0, 1, 2$ leading to gradient ($0 \to 1$), curl ($1 \to 2$), and divergence ($2 \to 3$) operations, respectively. 

These sequences may represent different (mechanical, electrical, thermal, etc.) domains of physics. Although, for most known physics, each domain's theory appears as one pair of (primary and secondary) sequences in tandem, connected by horizontal (or horizontal-diagonal) constitutive relations leading to Tonti diagrams, we do not make any such restriction when looking for new theories. The cross-sequence links can thus represent both single-physics constitutive relations and mutli-physics coupling interactions. Conservation laws, on the other hand, are represented by a balance between the output of a topological operator and an external source/sink, the latter being represented by a loop.

It is often more convenient to define product spaces (e.g., separate 3D space and 1D time, as opposed to 4D spacetime) in which conservation laws are stated as sums of incoming topological relations being balanced against an external source/sink. To accommodate such representations, we define abstract (symbolic) \inet{s} on a product of a $\Dimm_1-$space and a $\Dimm_2-$space as multi-sequences of co-chains, connected by phenomenological links, as before. It is possible to form $2^2 = 4$ possible such multi-sequences with various orientation combinations, two of which lead to so-called mechanical and field theories \cite{Tonti2013mathematical}, shown in Fig. \ref{fig_types} for (3+1)D spacetime and repeated in Fig. \ref{fig_inets} (b) for higher-dimensional pairs of abstract topological spaces. This construction is generalized to products of more than two spaces in a straightforward combinatorial fashion.


Based on the topological context, the semantics for co-boundary operators is unambiguously determined by the dimensions of the two variables (i.e., co-chains) they relate. However, phenomenological links require specifying a parameterization of possibly nonlinear, in-place, and purely metric relations they represent, using unknown parameters that must be learned from data.

\begin{figure*} [t]
	\centering
	\includegraphics[width=0.96\textwidth]{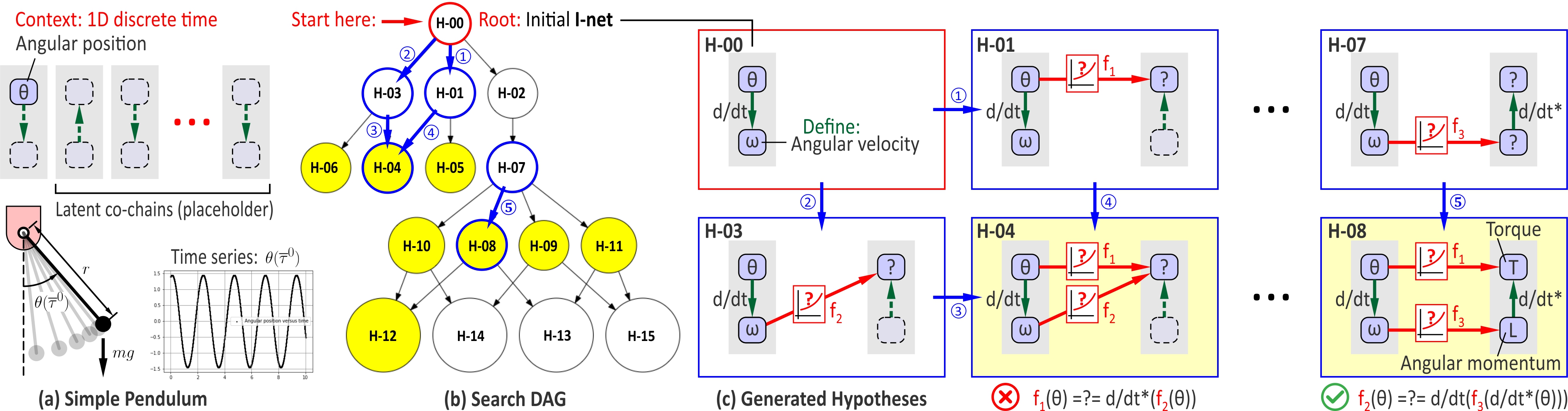}
	\caption{The search space for the dynamics of a pendulum in 1D time. The complete hypotheses (yellow nodes) correspond to \inet structures that pose new nontrivial equations to be tested against data, whereas incomplete hypotheses (white nodes) have ``dangling'' branches that are completed in their child states.} \label{fig_pend}
\end{figure*}


Once one or more hypotheses are specified in the language of abstract (symbolic) \inet{s} with unknown phenomenological parameters (e.g., thermal conductivity in the earlier heat transfer example), the parameters can be optimized to fit the data and the regression error can be used to evaluate the fitness of hypotheses.

\subsection{A Search for Viable Hypotheses}

Having defined a combinatorial representation of viable hypotheses that are partially ordered in terms of complexity, the next step is to generate and test the hypotheses in a ``simple-first'' fashion. 
The search space is defined by a directed acyclic graph (DAG) whose nodes (i.e., `states') represent symbolic \inet instances. The edges (i.e., state transitions) represent generating a new \inet structure by incrementally adding complexity to the parent state. Each action can be one or composition of (a) defining a new symbolic variable, in an existing co-chain complex, by applying a topological operator to an existing variable; (b) defining a new variable in a latent co-chain complex; and (c) adding phenomenological links of prescribed form and unknown parameters, connecting existing variables.
The search is guided by a loss function determined by how well the hypotheses represented by these \inet structures explain a given dataset. The algorithm may also be equipped with user-specified heuristic rules to prune the search space or prioritize paths that are perceived as ``more likely'' due to structural analogies with existing theories.

\begin{figure*} [t]
	\centering
	\includegraphics[width=0.96\textwidth]{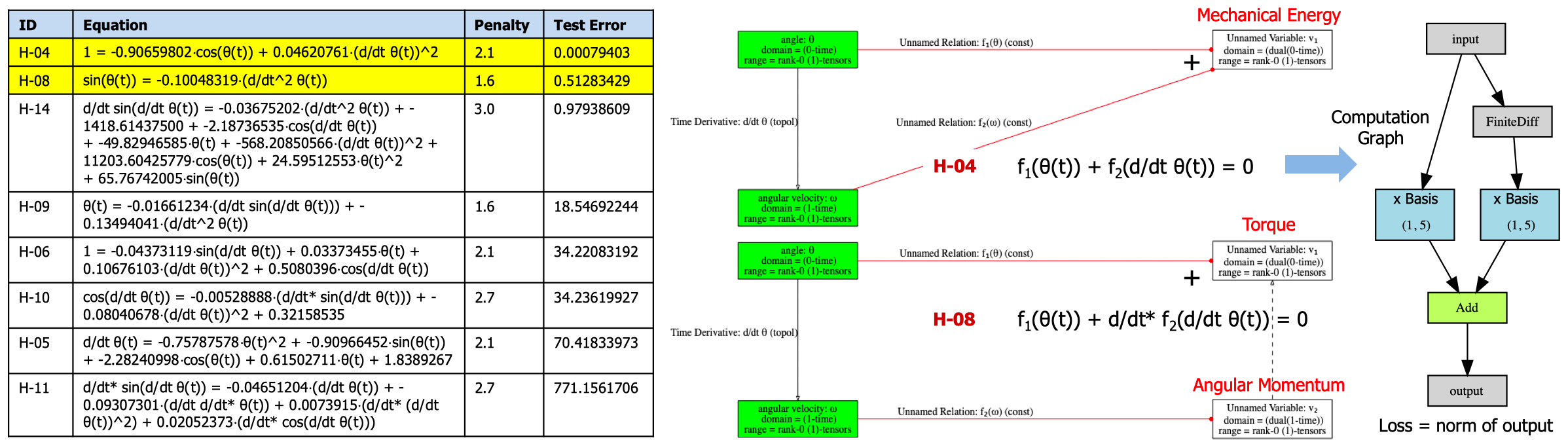}
	\caption{The hypotheses \textsf{H-04} and \textsf{H-08} of Fig. \ref{fig_pend} are enumerated and visualized by the software and evaluated against data (split 0.7-0.3 for training/testing). Both energy (first-order) and and torque (second-order) forms of the governing equation are discovered without human intervention. The former was quite unexpected, since its \inet structure does not correspond to a Tonti diagram. The latter has a larger error due to finite difference discretization.} \label{fig_pend_results}
\end{figure*}

The input to the search algorithm includes the bare minimum contextual information such as the assumed underlying topology, a preset number of physical domains, and the types of measured variables, e.g., spatiotemporal associations, tensor ranks and shapes, and dimensions/units. The search starts from an ``initial'' \inet instance (i.e., the `root') that embodies only measured variable(s) with no initial edges except the ones that are asserted a priori, e.g., loops for initial/boundary conditions or source terms, if applicable. The spatio-temporal types and physical semantics for these variables are provided by the experimentalist.

For example, consider a simple pendulum (Fig. \ref{fig_pend} (a)). We have only 1D time, leading to a topological space of inter-connected time instants $\overline{\tau}^0, \widetilde{\tau}^0 = \overline{\tau}^0 + \nicefrac{\epsilon}{2}$ and time intervals $\overline{\tau}^1 = (\overline{\tau}^0, \overline{\tau}^0+\epsilon), \widetilde{\tau}^1 = (\widetilde{\tau}^1, \widetilde{\tau}^1+\epsilon)$ to which data may be associated. Suppose we are given time series data for angular position $\theta(\overline{\tau}^0)$. The initial \inet instance is a single symbolic variable for this $0-$form, which can be differentiated only once in primary 1D time to obtain angular velocity as a $1-$form: $\theta(\overline{\tau}^0) \to \omega(\overline{\tau}^1) = \delta [\theta](\overline{\tau}^1)$ at the root of the search DAG (Fig. \ref{fig_pend} (b)). The DAG is expanded by adding new phenomenological links, either between two existing variables, or between an existing variable and one in a newly added latent co-chain sequence (Fig. \ref{fig_pend} (c)). In this example, the hypotheses are numbered \textsf{H-00} (the root) through \textsf{H-15}, enumerating all possible \inet structures formed by at most one latent co-chain complex in 1D time. The user can specify the maximum number of latent variables that the algorithm may consider, to keep the search tractable.

Not every introduction of new variables or relations makes nontrivial statements about physics.
%
%
For example, the hypothesis \textsf{H-01} produces a new variable typed as a $1-$pseudo-form $T(\widetilde{\tau}^1) = \ff_1(\theta(\dual\widetilde{\tau}^1))$, where the $\dual-$operator takes $\widetilde{\tau}^1$ to its dual: $\dual(\widetilde{\tau}^0, \widetilde{\tau}^0 + \epsilon) = \widetilde{\tau}^0 + \nicefrac{\epsilon}{2}$. However, until this new variable is reached through another path to close a cycle and pose a nontrivial equation, we do not have a complete hypothesis to (in)validate against data.
Further down the search DAG, \textsf{H-08} defines a new variable typed as a $0-$pseudo-form $L(\widetilde{\tau}^0) = \ff_2(\omega(\dual\widetilde{\tau}^0))$ where $\dual \widetilde{\tau}^0 = (\widetilde{\tau}^0 - \nicefrac{\epsilon}{2}, \widetilde{\tau}^0 + \nicefrac{\epsilon}{2})$. The co-boundary operation $L(\widetilde{\tau}^0) \to T(\widetilde{\tau}^1) = \delta [L](\widetilde{\tau}^1)$, closes the cycle and produces a commutative diagram (Fig. \ref{fig_pend} (c)) leading to:
\begin{equation}
	\mathcal{E}_{\textsf{H-08}}^{}(\theta; \ff_1, \ff_2) = \ff_1 (\theta) - \delta^\ast [\ff_2 (\delta[\theta])] = 0, \label{eq_pend}
\end{equation}
where $\ff_1, \ff_2$ are selected from restricted function spaces $\mathcal{F}_1, \mathcal{F}_2$ to avoid overfitting (e.g., parameterized by a linear combination of domain-aware basis functions) and their parameters must be determined from data to minimize the residual error $\mathcal{E}_{\textsf{H-08}}^{}$ over the entire period of data collection. A loss function can, for example, be defined as a mean-squared-error (MSE) to penalize violations uniformly over the time series period:
\begin{equation}
	\loss_{\textsf{H-08}}^{} = \min_{\ff_1 \in \mathcal{F}_1} \min_{\ff_2 \in \mathcal{F}_2} \big\| \mathcal{E}_{\textsf{H-08}}^{}(\theta; \ff_1, \ff_2) \big\|_{\widetilde{\tau}^1}, \label{eq_pend_loss}
\end{equation}
where $\|\cdot\|_{\widetilde{\tau}^1}$ is an $L_2-$norm computed as a temporal integral, i.e., sum of squared errors $\mathcal{E}_{\textsf{H-08}}^{2}(\theta; \ff_1, \ff_2)$ over time intervals $\widetilde{\tau}^1$ where \eq{eq_pend} is evaluated. In this example, it turns out that the best fit is achieved with $\ff_1(\theta) = \cc_1 \sin \theta$ and $\ff_2(\omega) = \cc_2 \omega$ where $\nicefrac{\cc_2}{\cc_1} = -\nicefrac{g}{r}$. The latent variables $L(\widetilde{\tau}^0)$ and $L(\widetilde{\tau}^0)$ turn out to be the familiar notions of angular momentum and torque, respectively, although the software need not know anything about them to generate and test what-if scenarios about their existence and correlations with angular position and velocity. Hence, {\it interpretability} of the discovered relationships by a human scientist does not require predisposing the AI associate to such interpretations, enabling unexpected discoveries.

\com{
In general, \inet instances can be classified as:
\begin{itemize}
	\item `reducible' \inet structures, in which removing any node and all of its incoming/outgoing eliminates an independent constraint in terms of other nodes; and
	\item `irreducible' \inet structures that are not reducible.
\end{itemize}
Every reducible structure can be mapped to a unique irreducible structure. The former can be pictured by \inet structures that have ``dangling'' branches that carry no new (and nontrivial) information.
The incremental steps in the search produces intermediate \inet structures that have such branches. Every time a dangling branch is turned into one or more closed cycles by adding enough new edges and (potentially) new nodes introducing latent variables, a new constraint is hypothesized that can be evaluated against data. 

The loss function for an irreducible \inet structure can be computed as a sum of residual errors for each of the independent constraints hypothesized by converging paths.

Note that reducible \inet hypotheses need not be evaluated separately, as their loss is the same as the corresponding irreducible structure, evaluated earlier at a parent state in the search tree.
}

In general, every state in the search DAG can be classified as complete or incomplete hypotheses. The former are \inet structures with ``dangling'' branches that carry no new nontrivial information in addition to their parent states. Every time such a branch is turned into one or more closed cycles by adding enough new variables and/or relations, a new constraint is hypothesized that can be evaluated against data.
%
%
%
When adding new dangling branches to the \inet structure, the search algorithm prioritizes actions that produce \inet structures similar to existing Tonti diagrams by assigning a penalty factor to every violation of the common structure (e.g., diagonal phenomenological links connecting non-dual cells). The loss for complete hypotheses can be computed as the sum of penalties for the \inet structure and the sum of residual errors for each of the independent constraints, implied by converging paths, multiplied by use-specified relative weight of the penalties and errors.  We use an A* algorithm to search the space of hypotheses. Since we cannot compute the error for incomplete hypotheses, we can only prune them when the increase in their penalty is large enough that it would fail even if it had no error at all.

\subsection{Generating Symbolic Expressions}

One of the practical features of our implementation in \python is its ability to automatically convert \inet instances to symbolic DE expressions in \sympy, when the co-boundary operators are interpreted in a differential setting for infinitesimal cells ($\epsilon \to 0^+$); for example, equation \eq{eq_pend} can be rewritten as a nonlinear ODE:
\begin{equation}
	\mathcal{E}_{\textsf{H-08}}^{}(\theta; \ff_1, \ff_2) = \ff_1 (\theta) - \frac{\partial}{\partial t} \left[\ff_2 \left(\dot{\theta}(t) \right) \right]. \label{eq_pend_ode}
\end{equation}
As a result, the generated hypotheses can be evaluated using any number of existing ML or symbolic regression frameworks that standardize on ODE/PDE inputs. 
For example, using non-orthogonal basis functions $\{1, x, x^2, \sin x, \cos x\}$ to span both function spaces $\mathcal{F}_1, \mathcal{F}_2$, we can substitute for both symbolic functions:
\begin{align}
	\ff_1(\theta) &:= \cc_0^1 + \cc_1^1 \theta + \cc_2^1 \theta^2 + \cc_3^1 \sin \theta + \cc_3^1 \cos \theta, \\
	\ff_2(\dot{\theta}) &:= \cc_0^2 + \cc_1^2 \dot{\theta} + \cc_2^2 \dot{\theta}^2 + \cc_3^2 \sin \dot{\theta} + \cc_3^2 \cos \dot{\theta},
\end{align}
into \eq{eq_pend_ode} to obtain a symbolic second-order (non)linear ODE in \sympy. Next, the software performs algebraic simplification to identify {\it equivalence classes} of hypotheses that, despite coming from different \inet structures, lead to the same ODE upon differential interpretation of the \inet{s}. For ODEs which, after simplification, are linear combinations of nonlinear (differential/algebraic) terms that are computable from data, we can apply symbolic regression to estimate the coefficients from data; for example, we tried LASSO-regularized least-squares regression in \pdefind \cite{Rudy2017data} where each term involving a derivative is evaluated using finite difference or polynomial approximation, whose results are shown in Fig. \ref{fig_pend_results}.

There are at least two issues with this approach:

First, more sophisticated regression or nonlinear programming methods are needed if the DE has terms that have nested nonlinear functions, i.e., cannot be represented as a linear combination of nonlinear terms because of unknown coefficients embedded within each term. We solve this problem by directly mapping \inet structures to computation graphs in \pytorch, skipping differential interpretation to symbolic DEs altogether. 

Second, numerical approximation of symbolic PDEs is a tricky business, as the discrete forms (in 3D space) may not obey the conservation principles postulated by the \inet structure after such approximations. It is difficult to separate discretization errors from modeling errors and noise in data. One of the key advantages of \inet{s} is the rich geometric information in their type system that is fundamental to physics-compatible and mimetic discretization schemes \cite{Koren2014physics,Palha2014physics,Lipnikov2014mimetic} that ensure conservation laws are satisfied {\it exactly} as a discrete level, regardless of spatial mesh or time-step resolutions. Such information is lost upon conversion to symbolic DEs. Retaining this information is even more important when dealing with noisy data, because discrete differentiation of noisy data (e.g., via finite difference or polynomial fitting) can substantially amplify the noise.

The good news is that we can {\it directly} interpret the same \inet instance in integral form to generate equations over larger regions in space and/or time, to make the computations more resilient to noise. For example, in the heat equation, the discrete divergence of heat flux over a single $3-$cell is replaced by a flux integral over a collection of $3-$cells, and is equated against the volumetric intgeral of internal energy within the collection. The cancellation of internal surface fluxes (discrete form of Gauss' divergence theorem) is built into the interpretation based on cellular homology. The integrals can be computed using higher-order integration schemes, e.g., using polynomial interpolation with underfitting to filter the noise.

Further details on directly and automatically mapping the abstract (symbolic) \inet structures to discrete (cellular) and numerical (tensor-based) \inet instance (e.g., computation graphs in \pytorch), learning scale-aware phenomenological relations, and physics-compatible discretization and de-noising will be presented in a full paper.

\begin{figure*} [t]
	\centering
	\includegraphics[width=0.96\textwidth]{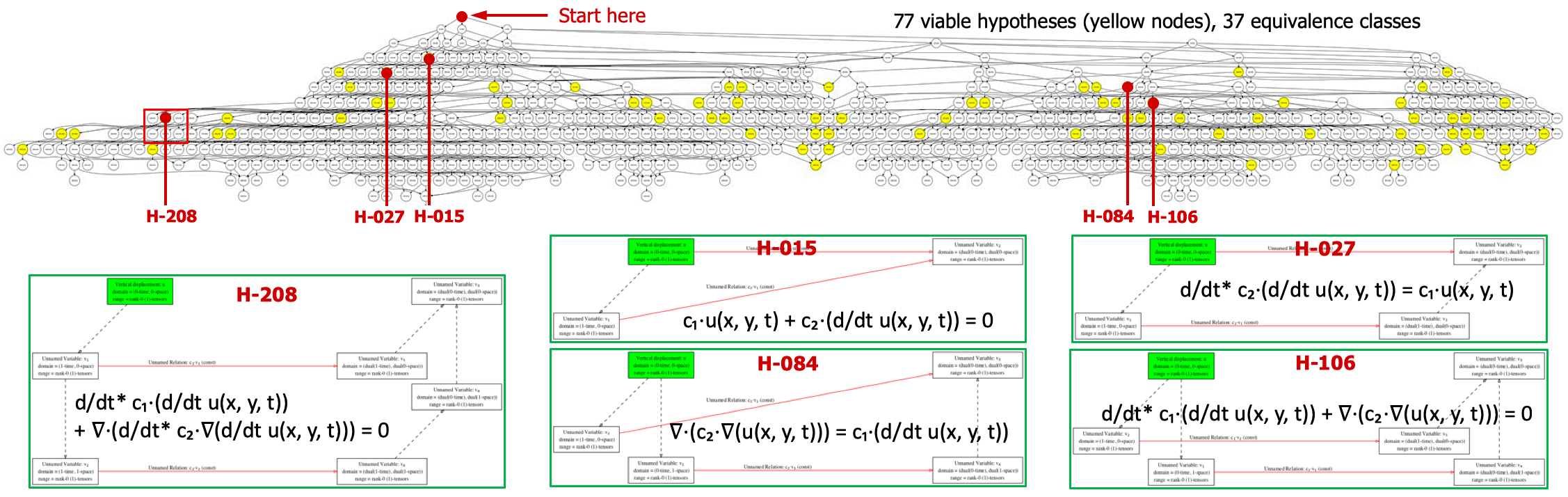}
	\caption{The search DAG and a number of viable hypotheses to explain ultrasound wavefield in metal parts.} \label{fig_afrl_dag}
\end{figure*}

\begin{figure*} [t]
	\centering
	\includegraphics[width=0.96\textwidth]{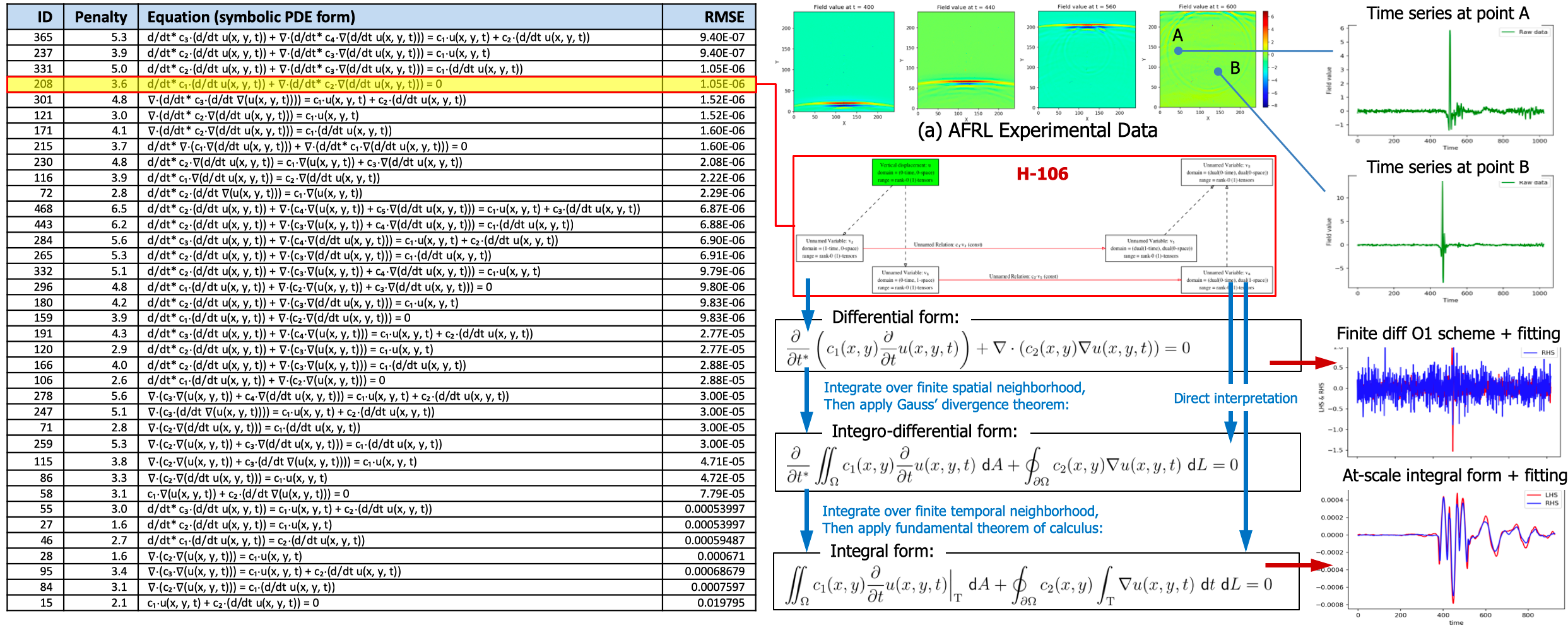}
	\caption{The AI associate discovers the (integral form) of the wave equation as well as the proper length/time scale at which the heterogeneous material properties (in this case, speed of sound) must be defined.} \label{fig_afrl_results}
\end{figure*}

\section{Real-World Scientific Discovery}

Figures \ref{fig_afrl_dag} and \ref{fig_afrl_results} illustrate the application of our AI approach to an elastodynamics challenge problem provided by AFRL in the course of the DARPA AI Research Associate (AIRA) program that supported the development of \cyphy. The input is noisy data obtained by ultrasound imaging, measured in (2+1)D spacetime over the surface of several material samples with heterogeneous properties. 

Figure \ref{fig_afrl_dag} illustrates the search DAG along with a number of \inet structures for viable hypotheses, each postulating the relevance of a conservation law and existence of a few phenomenological relations. 
Figure \ref{fig_afrl_results} shows the ranking of these hypotheses based on their residual errors when tested against data. Each hypothesis can be interpreted in differential, integral, or integro-differential forms. The results demonstrate that integral forms applied to wide spatial and temporal neighborhoods (of $\sim25$ grid elements along each axis) with high-order polynomial underfitting (up to cubic in each coordinate), resulting in a length/time scale-aware definition of (nonlocal) phenomenological relations as well as physics-compatible (i.e., mimetic) discretization and de-noising, are preferable to strictly local numerical schemes such as finite difference discretization.


\section{Conclusion}

Statistical learning methods, despite their accuracy and efficiency in narrow regimes for which they are carefully engineered, are not sufficient to independently acquire {\it deep understandings} of the scientific problems they are applied to. Human scientists continue to handle most of knowledge-centric aspects of the scientific process based on domain-specific insight, experience, and expertise.

Our novel approach to early-stage scientific hypothesis generation and testing demonstrates a path forward towards context-aware, generalizable, and interpretable AI for scientific discovery. Our AI associate (\cyphy) distinguishes between non-negotiable mathematical truism, implied by the relationship between measurement and presupposed spacetime topology, and phenomenological realities that are at the mercy of empirical learning. Data-driven regression is targeted at the latter to enable distilling governing equations from sparse and noisy data, while providing deep insights into the mathematical foundations.

\section{Acknowledgment}
This material is based upon work supported by the Defense Advanced Research Projects Agency (DARPA) under Agreement No. HR00111990029.
%

{\footnotesize
\bibliography{CyPhyShort}
}

\end{document}